\documentclass{article}

\usepackage{arxiv}

\usepackage[utf8]{inputenc} 
\usepackage[T1]{fontenc}    
\usepackage{hyperref}       
\usepackage{url}            
\usepackage{booktabs}       
\usepackage{amsfonts}       
\usepackage{nicefrac}       
\usepackage{microtype}      
\usepackage{lipsum}
\usepackage{svg}
\usepackage{graphicx}
\graphicspath{ {./images/} }
\usepackage{amsmath} 
\usepackage{pdflscape}
\usepackage{xcolor}
\usepackage[normalem]{ulem}
\usepackage{comment}
\usepackage{longtable}
\usepackage{tabularx}
\usepackage{caption} 
\usepackage{subcaption}
\usepackage{longtable}
\usepackage{booktabs}
\usepackage[utf8]{inputenc}
\usepackage{changepage}

\usepackage{array}
\usepackage{float}
\newlength{\extralength}
\newlength{\fulllength}
\newcolumntype{C}{>{\centering\arraybackslash}X}

\title{Performance of YOLOv7 in Kitchen Safety While Handling Knife}

\author{Athulya Sundaresan Geetha\\[1ex]
\begin{minipage}[t]{0.90\textwidth}
\centering
\scriptsize Department of Computer Science, Huddersfield University, Queensgate, Huddersfield HD1 3DH, UK; \\
\textsuperscript{*}Correspondence: U2282847@unimail.hud.ac.uk;
\end{minipage}}

\begin{document}

\maketitle
\begin{abstract}Safe knife practices in the kitchen significantly reduce the risk of cuts, injuries, and serious accidents during food preparation. Using YOLOv7, an advanced object detection model, this study focuses on identifying safety risks during knife handling, particularly improper finger placement and blade contact with hand. The model's performance was evaluated using metrics such as precision, recall, mAP50, and mAP50-95. The results demonstrate that YOLOv7 achieved its best performance at epoch 31, with a mAP50-95 score of 0.7879, precision of 0.9063, and recall of 0.7503. These findings highlight YOLOv7's potential to accurately detect knife-related hazards, promoting the development of improved kitchen safety.
\end{abstract}


\section{Introduction}
Knife safety is an important part of food preparation that is often overlooked despite its importance in preventing accidents and injuries. Improper knife handling can lead to severe cuts, injuries, and even long-term damage, making it essential for amateur and experienced cooks to understand proper safety practices. The key to maintaining knife safety involves careful attention to the positioning of hands and fingers and the knife handling: curling the fingers of the hand to grip items that need to be cut and hand making contact with the blade or tapping on the top of the blade. Ensuring that knives are used correctly not only improves efficiency but also minimizes the risk of accidents. This includes practices such as holding the knife securely and positioning fingers away from the blade. With the right safety measures, knife-related accidents can be significantly reduced, creating a safer kitchen environment.

Detecting hazards associated with safe knife handling presents several challenges. One of the major issues is the variation in lighting conditions within the kitchen, which can alter how the knife appears in an image, making it harder to detect accurately. Additionally, kitchens are often filled with appliances, utensils, and ingredients that can obstruct, either partially or completely, the knife from view. Sometimes poor image quality complicates the detection process. Furthermore, other kitchen tools may resemble a knife in shape or color, leading to misidentifications. Another challenge is the interpretation of hazards, such as improper knife handling, fingers extending over the cutting surface, or tapping on the blade, which can be difficult for models to recognize accurately.

Histogram of oriented gradients (HOG) focused on horizontal and vertical features, and vertical histograms oriented gradients (VHOG) concentrated vertical features \cite{RN1}. Support vector machine (SVM) and extreme learning machine (ELM)classified movements but struggled with accuracy \cite{RN2, RN3, RN4}. Due to the time-consuming nature and frequent need for manual supervision, CNN models were developed. For detecting objects, combining Faster R-CNN and GoogleNet demonstrated superior performance \cite{RN5}. Various models, including ResNet \cite{RN8}, Google-Net \cite{RN7}, Region-based Convolutional Network method (R-CNN) \cite{RN10}, AlexNet \cite{RN6}, Fast-RCNN \cite{RN11}, Faster-RCNN \cite{RN12}, and VGG-Net \cite{RN9} were developed to advance object identification techniques \cite{RN13}.

You Only Look Once (YOLO) models effectively resolve limitations in data processing, speed, and architecture,  issues that relate to detection accuracy in two-stage detection processes. With DarkNet at its core, YOLOv1 includes 24 convolutional layers, while Fast YOLO has 9 \cite{RN14}. YOLOv2, an enhancement over previous models, relies on DarkNet-19 and features clustering focused on size, normalizing input batches, referencing through box, depth class identification, finer details \cite{RN15}. YOLOv3, built on DarkNet-53, incorporates residual networks for feature extraction and uses binary cross-entropy as its loss function \cite{RN16}.In YOLOv4, CSPDarkNet53 serves as the backbone architecture, and PANet with SPP functions as the neck network for improved performance \cite{RN17}.

CSP-PAN as the neck and SPPF as the head contribute to YOLOv5’s exceptional performance speed \cite{RN18}. YOLOv6 not only enhances performance speed but also reduces the complexity of computational demands \cite{RN19}. YOLOv7 achieves greater accuracy in detecting objects through the integration of a lead head and an auxiliary head \cite{RN20}. Identification of objects and segmentation of images are achieved in YOLOv8 through a revamped CSPDarkNet53 backbone and PAN-FPN neck. Object detection accuracy was improved in YOLOv9 and YOLOv10 through Generalized Efficient Layer Aggregation Network and Programmable Gradient Information, with YOLOv10 excluding non-maximum suppression \cite{RN21, RN22}.

This study evaluates the performance of YOLOv7 in identifying hazards associated with knife safety within a kitchen environment. Key metrics such as accuracy, recall, mAP50, mAP50-95, F1 score, and the confusion matrix are analyzed to assess its effectiveness. The findings aim to enhance real-time safety systems by providing alerts when knives are handled improperly, following established safety guidelines to prevent accidents.

The structure of the paper is outlined as follows: The Literature Review section examines knife handling studies with a special emphasis on YOLO models. In the Methodology section, a comprehensive explanation of YOLOv7's workflow and architectures is provided. The Experimental Results and Discussion sections explore the performance matrices in detail.

\section{Literature Review} 

Safety in knife handling depends on the ability to recognize threats and take measures to avoid accidents. Although support vector machines analyzed local visual and motion features for offline classification \cite{RN2, RN3}, they were unable to detect food preparation tasks \cite{RN4}. To classify actions like slicing into cubes, stripping off the peel, blending thoroughly, and scooping out, data from sensors were recognized by analyzing metrics such as mean score, power, statistical variance, randomness factor, and spatial alignment
\cite{RN23}. A different study improved activity recognition by incorporating Wii controllers and combining their sensor data with an RGBD camera positioned to monitor the kitchen area \cite{RN24}. Haar filters were used with computer vision to detect knives, processing images based on dimension, arrangement, and structure, achieving a 45\% correct detection rate and 85\% misclassification rate \cite{RN25}. Drawbacks encompass longer processing times, high-level architectures, and the necessity for constant human supervision. The use of Convolutional Neural Networks (CNNs) led to substantial improvements in computer vision applications, particularly in image analysis \cite{RN8, hussain2023custom, RN10, RN11, RN12, RN9, hussain2022feature, RN26}. Faster R-CNN obtained 46.68\% accuracy for knife detection, while SqueezeNet achieved a higher accuracy of 85.44\% for gun detection \cite{RN5}. Retinex and K-means++ helped Faster R-CNN achieve a mAP of 94.3\% in helmet detection, with a speed of 11.62 images per second \cite{RN40}. Using MobileNet for classification with 95\% accuracy, MaskRCNN for detection and segmentation, and PoseNet for skeletal point positioning, the knife and threat detectors were improved \cite{RN27}.

YOLO demonstrated 97.4\% accuracy for gun detection, and ResNet achieved 73.2\% accuracy for knife detection in the study's X-ray baggage security framework \cite{RN28}. YOLOv2 detected objects for visually impaired individuals in cluttered environments, combining with shape-oriented methods to estimate three-dimensional models in kitchens, offering safety instructions at 90.45\% accuracy and 0.86 frames per second \cite{RN29}. Using YOLOv3, VGG-16 achieved a higher Average Precision (AP) of 62.2\% for hand detection, outperforming MobileNet-Lite, although dynamic hand movements and noise were limitations \cite{RN30}. For real-time safety management in on-site power work, a lightweight YOLOv4 model was proposed using depthwise separable convolutions, a mobile-inverted bottleneck structure, and an enhanced bidirectional fusion network, achieving a 93.11\% parameter reduction, a 22\% detection speed boost, and 84.82\% accuracy \cite{RN31}.

SafeCOOK, a system using YOLOv5 and Kernelized Correlation Filter, tracked kitchen appliances to enhance safety by identifying environment. YOLOv5 struggled with overlapping objects, while combining with Single and Multiple Object Tracking reduced errors. The model achieved high performance with mAP50 of 97.6\%, 89 epochs, and 10 frames per second \cite{RN32, RN33, RN34, RN35}. YOLOv6 was utilized for fire detection in smart cities, achieving 98\% performance, 96\% recall, and 83\% precision, and will be further developed for better handling of difficult scenarios and a three-dimensional CNN/U-Net approach \cite{RN36}. Using YOLOv7 and YOLOv8 architectures, weapon detection was performed with accurate bounding box annotations, where YOLOv7-e6 achieved a 90.3\% mAP at a 0.5 IoU threshold, and future developments will enhance YOLOv8 for real-time object detection\cite{RN37}.

In a study using YOLOv8, a pre-trained model was tested for knife detection, and a custom model was trained for knives \cite{RN38}, while another study employed YOLOv8 for detecting violations in hygiene with 89\% accuracy for food safety \cite{RN39}. The study introduces Ghost Convolution (GC)-YOLOv9 for real-time traffic surveillance, achieving mAP0.5 scores of 77.15 and 74.95 on the BDD100K and Cityscapes datasets, respectively, by enhancing smart city applications in safety protocols, fire detection, and security networks \cite{RN200}.  In a study by Athulya and Hussain, YOLOv5, YOLOv8, and YOLOv10 for knife safety detection across five classes were compared, YOLOv5 and YOLOv8 performed well in identifying hazard 1 and hazard 2, respectively, and all three YOLO models performed equally in detecting cutting board \cite{geetha2024comparative, RN204, RN205, RN206}.

This literature review focuses on the application of object detection, specifically YOLO models. YOLO models show significant advancements in optimization, acceleration, and scalability, handling blurred images and recognizing small or hard-to-detect objects efficiently. In this study, securely holding vegetables with curled fingers and tapping the blade are two hazards that have been considered and analyzed for the performance of YOLOv7.

\section{Methodology}

\subsection{Dataset}
\begin{figure}[H]
\begin{adjustwidth}{-\extralength}{0cm}
\centering
\includegraphics[width=15cm]{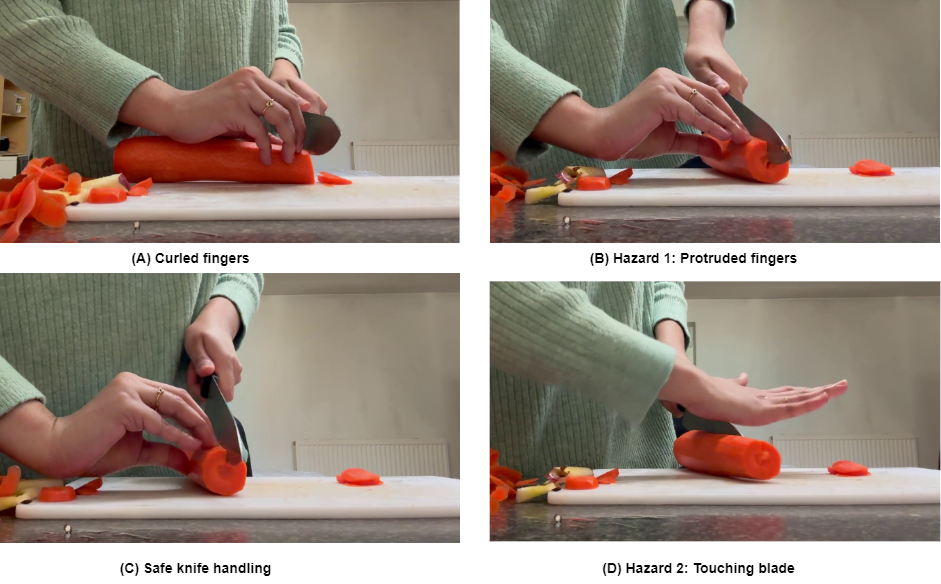}
\end{adjustwidth}
\caption{Knife hazards sample.}
\label{Figure:1}
\end{figure}

The dataset used in this study was captured with an Apple iPhone15 Pro, where the video's resolution is 1920 × 1080 pixel. The video underwent processing with a Python script, resulting in 6,004 individual frames. Manually processed in Label Studio, the frames were sorted into six categories: cutting board, hands, vegetable, knife, hazard 1 (curl finger), and hazard 2 (hand touching blade). The investigation of safe knife use in the kitchen identified hazard 1 as curling fingers to avoid accidental injuries, and hazard 2 as a hand making contact with the blade to prevent serious harm. Figures \ref{Figure:1}A-D show that the hand-holding vegetable is curled (Figure \ref{Figure:1}A) and fingers are extended (Figure \ref{Figure:1}B), as well as safe handling of knife (Figure \ref{Figure:1}C) and hand touching the blade (Figure \ref{Figure:1}D).

\subsection{Data Augmentation}
Data preprocessing enhances image analysis in machine learning and computer vision by increasing dataset diversity used in various applications such as pallet racking ~\cite{hussain2022gradient}, Medical ~\cite{aydin2023domain} and emotion detection ~\cite{hussain2023child} applications . It addresses challenges such as class imbalance, overfitting, and varying conditions, like lighting and occlusions, essential for real-world applications. By introducing transformations like rotations, scaling, and variations in lighting, the model becomes robust, adapting to unseen data and diverse scenarios, such as detecting kitchen hazards with dynamic hand gestures and obstructed views. This improves model accuracy, ensuring safer knife-handling practices in kitchen environments.

\subsubsection{Flip}
A model trained on flipped and unflipped images will be more helpful in detecting objects or actions regardless of positioning. The horizontal flip option flips the image along its vertical axis (left-to-right). The vertical flip option flips the image along its horizontal axis (top-to-bottom).

Equation \ref{eq:1} given below is for horizontal flipping of the image.

\begin{equation}
    F^{\prime}(x, y) = F(W- x - 1, y),
    \label{eq:1}
\end{equation}

where \ \(F^{\prime}\) is the resulting image; \(F\) is the original image; W indicates width; and x and y mean horizontal and vertical coordinates, respectively.

\subsubsection{Crop}
Cropping helps in centering the key features of the image, such as objects or actions, which are crucial for model training. Here, the minimum cropped image is 0\%, i.e., original image, and the maximum zoom is 20\%, where 20\% of the image’s width and height is removed.

Equation \ref{eq:2} denotes 20\% crop at width and height of the images.
 
\begin{equation}
    F^{\prime}=F[ top +p \times height, bottom -p \times height, left +p \times width, right -p
    \label{eq:2}
\end{equation}

\(p\) indicates 20\% cropping.

\subsubsection{Rotation}
By generating rotated images at different angles, the model generalizes better to new and unseen data with different positions. When the image is rotated counterclockwise by 15 degrees (-15 degrees), the subject appears tilted to the left. When the image is rotated clockwise by 15 degrees (+15 degrees), the subject appears tilted to the right.

Equation \ref{eq:3} shows the mathematical representation for rotation.

\begin{equation}
    F^\prime\ =\ rotate(F,\theta)
    \label{eq:3}
\end{equation}

\subsubsection{Random Grayscale}
\begin{figure}[H]
\begin{adjustwidth}{-\extralength}{0cm}
\centering
\includegraphics[width=12cm]{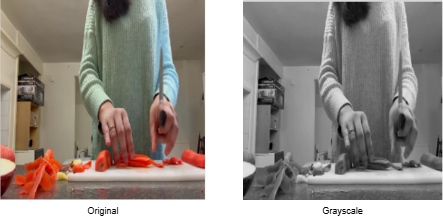}
\end{adjustwidth}
\caption{Grayscle image}
\label{Figure:2}
\end{figure}

Processing 15\% of colored images in grayscale strengthens the model’s performance by emphasizing textures and shapes, reducing dependence on colors. The intensity of light (brightness) is preserved, ranging from black to white (Figure \ref{Figure:2}).

Equation \ref{eq:4} shows the formula for grayscale.

\begin{equation}
    F' = 
    \begin{cases} 
        0.299 \times R + 0.587 \times G + 0.114 \times B, & \text{with probability } p \\
        F, & \text{with probability } (1-p)
    \end{cases}
    \label{eq:4}
\end{equation}

\(R\) denotes red, \(G\) means green, and \(B\) indicates blue; \(p\) represents the transformation probability.

\subsubsection{Shear}
Shear preprocessing technique distorts the image along one axis to simulate changes in perspective. The images are horizontally sheared by +10 degrees and -10 degrees, respectively, creating a slanted or tilted effect along the horizontal axis, where objects appear to lean left or right. The images are vertically sheared by +10 degrees and -10 degrees, respectively. This creates distortion along the vertical axis, where objects appear stretched or compressed vertically.

For distortion, shearing allows the object or subject to appear to be slanted (Equation \ref{eq:5}).

\begin{equation}
    F^\prime\ =\ F\cdot S
    \label{eq:5}
\end{equation}

\textit{S} means shear matrix.

\subsubsection{Hue}
Hue augmentation adjusts the colors of an image. Shifting the hue by -25 degrees and +25 degrees causes a color change where the overall tone is shifted to a different part of the color spectrum.

Equation \ref{eq:6} shows the notation for hue.

\begin{equation}
    F^\prime=F+H
    \label{eq:6}
\end{equation}

\subsubsection{Saturation}
Saturation augmentation randomly adjusts the vibrancy of the colors in an image, either increasing or decreasing it. This makes the model less sensitive to changes in color intensity, improving its robustness and generalization to various real-world conditions. The result of reducing the saturation by 25\%, making the image less vibrant; colors appear duller and closer to grayscale but still retain some color information. The result of increasing the saturation by 25\%, making the colors more vibrant and intense.

The saturation formula is given in Equation \ref{eq:7}.

\begin{equation}
    F^\prime\ =\text{adjust\_saturation}(F,\alpha)
    \label{eq:7}
\end{equation}

The saturation adjustment factor is identified by $\alpha$.

\subsubsection{Brightness}
Brightness adjusts the light intensity of an image, either brightening or darkening it. This helps the model to adapt to various brightness levels to detect hazards correctly in different environments. The brightness variation is set to 15\%, meaning the images can randomly be brightened or darkened by up to 15\% during augmentation.

Equation \ref{eq:8} indicates notation of brightness. An adjusting factor is given as \ \(\beta\).

\begin{equation}
    F^\prime\ =\ clip(F\ +\ \beta\ \times255,\ 0,255)
    \label{eq:8}
\end{equation}

\subsubsection{Exposure}
Exposure augmentation adjusts the overall brightness of an image, either darkening or brightening it. The image's exposure has been reduced by 10\%, making it appear darker. This displays underexposed conditions, such as low light. The image's exposure has been increased by 10\%, making it appear brighter. This indicates overexposed conditions, such as direct sunlight.

Equation \ref{eq:9} denotes expression for exposure, with \(A\) an exposure adjustment factor.

\begin{equation}
    F^\prime=F\times\left(1+\frac{A}{100}\right),
    \label{eq:9}
\end{equation}

\subsubsection{Blur}
Blur augmentation involves applying Gaussian blur to an image to reduce sharpness and simulate out-of-focus conditions. The result shows that when applying a 2.5px Gaussian blur, the image becomes slightly blurred. The sharpness of edges and fine details in the image is reduced, mimicking a scenario where the subject is slightly out of focus.

$\sigma$ indicates the standard deviation of the Gaussian blur kernel (Equation \ref{eq:10}).

\begin{equation}
    F^\prime\ =\ Blur(F,\sigma),
    \label{eq:10}
\end{equation}

\subsubsection{Noise}
Noise augmentation simulates real-world imperfections by introducing random noise to an image. The result of applying 1.01\% noise indicates that small random speckles are visible throughout the image, simulating common imperfections like graininess or static caused by low light or sensor errors.

Equation \ref{eq:11} is given below:

\begin{equation}
    F^\prime\ =\ F+N
    \label{eq:11}
\end{equation}

\subsubsection{Cutout}
Cutout augmentation randomly removes sections of an image, simulating object occlusion. This helps train model to be more resilient to missing or obscured data. About 10\% of the image's area is obscured by three black rectangles. These rectangles represent regions of the image that have been randomly removed to simulate occlusion.

Equation \ref{eq:12} is given as follows:

\begin{equation}
    F'(x, y) =
    \begin{cases} 
    0 & \text{if } (x, y) \in \text{cutout regions}, \\
    F(x, y) & \text{otherwise},
    \end{cases}
    \label{eq:12}
\end{equation}

where \( F(x, y) \): the pixel value at coordinates \( (x, y) \) in the original image; \( F'(x, y) \): the pixel value at coordinates \( (x, y) \) in the augmented image; and cutout regions: rectangular patches that are randomly placed on the image.

\subsection{Architecture of YOLOv7}

YOLOv7's performance across a broad range of FPS (5-160) highlights its adaptability for both high-speed and high-resolution requirements. At 56 FPS, YOLOv7-E6 outperforms SWIN-L Cascade-Mask R-CNN (9.2 FPS, 53.9\% AP) and ConvNeXt-XL Cascade-Mask R-CNN (8.6 FPS, 55.2\% AP), achieving 509\% and 551\% faster speeds, respectively, along with better accuracy. YOLOv7's approach avoids reliance on pre-trained weights or external datasets, being trained solely from scratch on the MS COCO dataset. YOLOv7 backbones skip ImageNet pre-trained backbones and instead utilize Extended Efficient Layer Aggregation Network (E-ELAN) as their computational block. While Network Architecture Search (NAS) performs independent, parameter-specific scaling, a compound model scaling approach, scaling width and depth coherently, further optimizes concatenation-based models \cite{RN41}. By using Bag of Freebies, models achieve enhanced performance without incurring additional training costs, while re-parameterization improves inference by replacing the 3×3 convolution layer in E-ELAN with RepConv, alongside experiments adjusting RepConv, 3×3 Conv, and identity connection (a 1×1 convolutional layer) \cite{RN201, RN202, RN203}.

In the architecture of a YOLO-based object detection framework, the backbone extracts multi-scale features from the input image through layers. Then, these features are input into the Feature Pyramid Network (FPN), which integrates multi-resolution data through techniques like upsampling and concatenation for better feature representation. Multiple detection heads process the resulting feature maps to output object classification, bounding box regression, and objectness score predictions. The architecture employs Conv Blocks for feature processing, Upsample Blocks for modifying spatial resolution, and Concatenate operations to combine features. Allow injection points in the architecture to substitute components like the 3×3 convolution with more efficient or specialized options. This architecture ensures efficient and accurate object detection across scales over other detection models (Table \ref{tab:yolov7_architecture}).

\begin{table}[H]
\caption{YOLOv7 Architecture.\label{tab:yolov7_architecture}}
\begin{adjustwidth}{-\extralength}{0cm}
    \begin{tabularx}{\fulllength}{CCCCCC}
        \toprule
        \textbf{Layer} & \textbf{Activation} & \textbf{Filters} & \textbf{Size} & \textbf{Repeat} & \textbf{Output Size} \\
        \toprule
        Image          & -          & -       & -          & -      & 640 × 640 \\
        Conv0          & SiLU       & 32      & 3 × 3 / 2  & 1      & 320 × 320 \\
        Conv1          & SiLU       & 64      & 3 × 3 / 2  & 1      & 160 × 160 \\
        Conv2          & SiLU       & 128     & 3 × 3 / 2  & 1      & 80 × 80  \\
        Conv3          & SiLU       & 256     & 3 × 3 / 2  & 1      & 40 × 40  \\
        ELAN0          & SiLU       & 256     & -          & 1      & 40 × 40  \\
        Conv4          & SiLU       & 512     & 3 × 3 / 2  & 1      & 20 × 20  \\
        ELAN1          & SiLU       & 512     & -          & 1      & 20 × 20  \\
        Conv5          & SiLU       & 1024    & 3 × 3 / 2  & 1      & 10 × 10  \\
        ELAN2          & SiLU       & 1024    & -          & 1      & 10 × 10  \\
        SPPCSPC        & SiLU       & 1024    & -          & 1      & 10 × 10  \\
        Conv6          & SiLU       & 512     & 1 × 1      & 1      & 20 × 20  \\
        Upsample       & -          & -       & -          & 1      & 20 × 20  \\
        C3             & SiLU       & 256     & -          & 1      & 20 × 20  \\
        Upsample       & -          & -       & -          & 1      & 40 × 40  \\
        C3             & SiLU       & 128     & -          & 1      & 40 × 40  \\
        Head           & SiLU       & Varies  & -          & -      & Varies   \\
        \bottomrule
    \end{tabularx}
\end{adjustwidth}
\end{table}

To detail, the network begins with the input image and applies a series of convolutional layers, which progressively extract features while reducing spatial resolution and increasing channel depth. These features are further processed through ELAN blocks, which efficiently aggregate multi-scale features using a combination of convolution, concatenation, and other operations. The Spatial Pyramid Pooling - Cross Stage Partial Connections (SPPCSPC) enhances the network's ability to detect objects of varying sizes by pooling multi-scale contextual information and fusing it into the feature map. he architecture also incorporates upsampling layers to increase spatial resolution, allowing multi-scale feature fusion through C3 blocks. Finally, the Head layer produces predictions for object detection, including class probabilities, bounding box coordinates, and confidence scores, ensuring robust detection at multiple scales. YOLOv7’s modular design, which combines downsampling, upsampling, and multi-scale fusion, enables high efficiency and accuracy, making it one of the most effective real-time object detection models, which makes YOLOv7 better than other previous YOLO models.

\section{Experimental Results}
The performance of YOLOv7 during training and validation is analyzed and discussed in this section. It achieved accuracy improvements by training and validating in a PyTorch environment with NVIDIA GPUs, fine-tuning hyperparameters over 40 epochs. Training employed the AdamW optimizer, set to a learning rate of 0.001 and a momentum factor of 0.9. Fine-tuning targeted parameter groups by excluding weights from decay adjustments and applying L2 regularization to biases with a decay rate of 0.0005. By mitigating overfitting, this method allowed the model to retain strong generalization performance.

\begin{figure}[H]
\begin{adjustwidth}{-\extralength}{0cm}
\centering
\includegraphics[height=8cm]{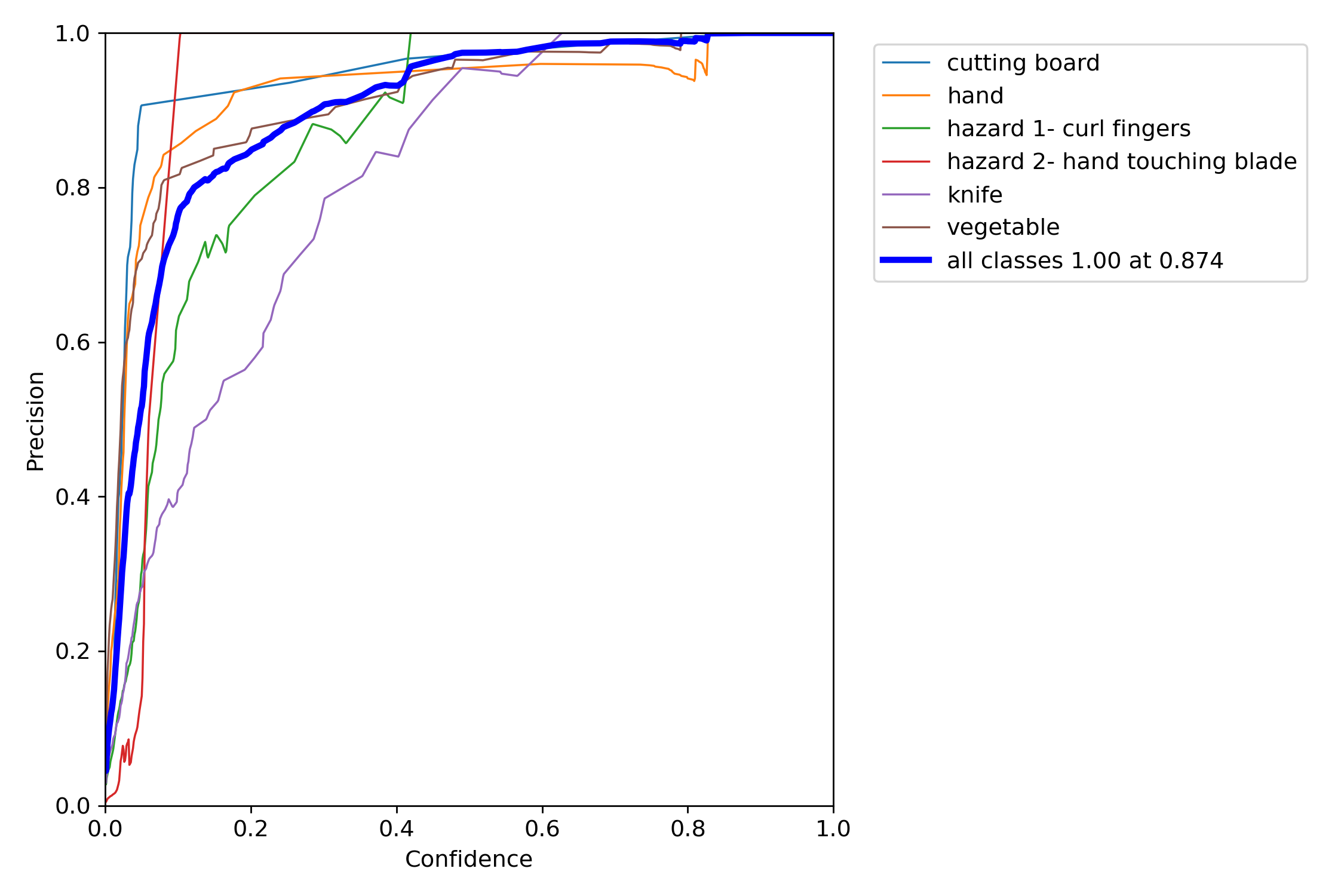}
\end{adjustwidth}
\caption{Precision-confidence value of YOLOv7 for Kitchen dataset.}
\label{Figure:3}
\end{figure}

A precision-confidence curve for an object detection model explains how precision varies across different confidence thresholds for each class (Figure \ref{Figure:3}). The x-axis denotes the confidence score thresholds, ranging from 0.0 to 1.0, while the y-axis indicates precision, the proportion of true positive predictions out of all positive predictions. The combined performance across all classes, with the note, 1.00 at 0.874, indicated that the model achieves perfect precision (1.00) for all classes at a confidence threshold of 0.874. Classes like "cutting board" demonstrate consistently high precision even at lower confidence thresholds, while "hazard 2 - hand touching blade" shows reduced precision at lower thresholds, indicating a higher occurrence of false positives

\begin{figure}[H]
\begin{adjustwidth}{-\extralength}{0cm}
\centering
\includegraphics[height=8cm]{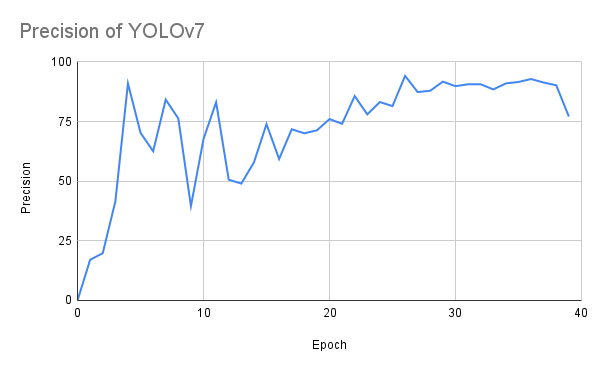}
\end{adjustwidth}
\caption{Precision of YOLOv7.}
\label{Figure:4}
\end{figure}

The graph depicts the precision of the YOLOv7 model across training epochs (Figure \ref{Figure:4}). The x-axis represents the number of epochs, ranging from 0 to 40, while the y-axis indicates precision as a percentage, from 0 to 100. The precision value fluctuates throughout the training process, starting at a low level during the initial epochs and rapidly increasing as the model learns. As training progresses, the precision stabilizes around the middle epochs, with occasional dips and peaks reflecting the model's adaptation to the dataset. Toward the later epochs, the precision consistently raises near higher values, indicating that the model has improved its ability to make accurate predictions. However, a slight decline at the end suggests either overfitting or the need for further fine-tuning. Overall, the graph highlights the training dynamics and the model's gradual improvement in precision over time.

\begin{figure}[H]
\begin{adjustwidth}{-\extralength}{0cm}
\centering
\includegraphics[height=8cm]{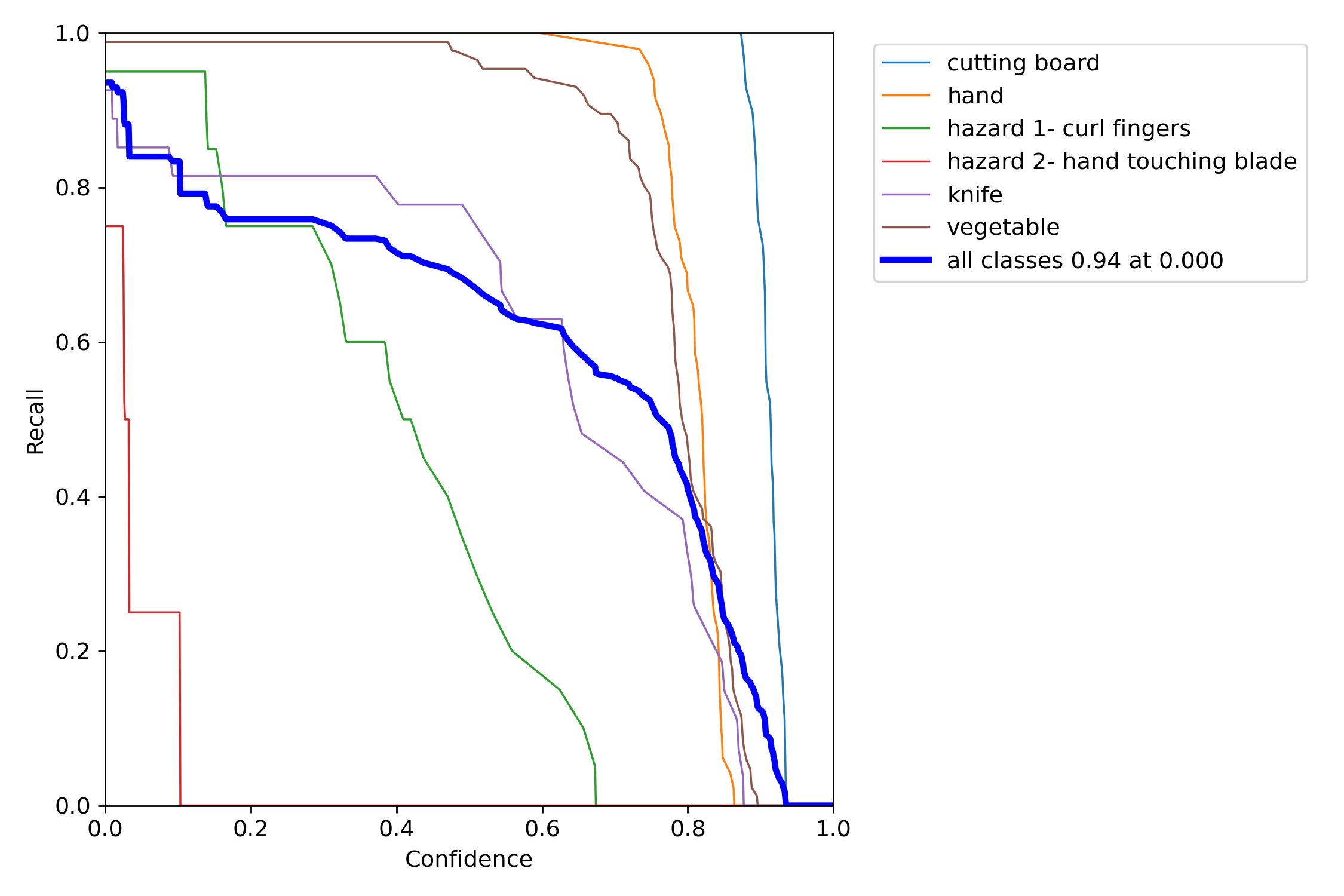}
\end{adjustwidth}
\caption{Recall-confidence value of YOLOv7.}
\label{Figure:5}
\end{figure}

In Figure \ref{Figure:5}, the x-axis represents the confidence score thresholds, ranging from 0.0 to 1.0, while the y-axis displays recall, which measures the proportion of correctly detected objects. The combined recall performance across all classes, with a maximum recall of 0.94, achieved a confidence threshold of 0.000. Certain classes, such as "cutting board" and "hand," maintain high recall even at higher confidence thresholds, indicating the model’s strong detection ability for these categories.

\begin{figure}[H]
\begin{adjustwidth}{-\extralength}{0cm}
\centering
\includegraphics[height=8cm]{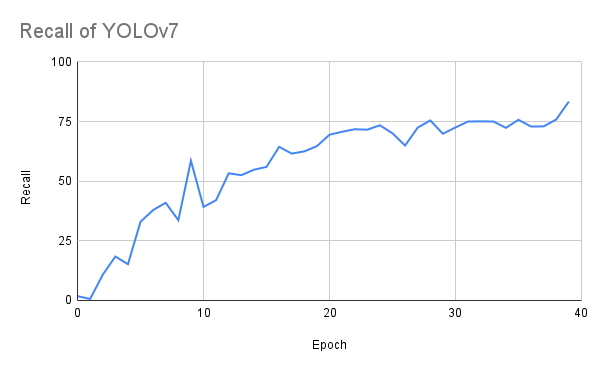}
\end{adjustwidth}
\caption{Recall of YOLOv7.}
\label{Figure:6}
\end{figure}

The graph depicts the recall of the YOLOv7 model across training epochs (Figure \ref{Figure:6}). The curve starts at a low value in the initial epochs, reflecting the model's limited ability to detect positive instances early in training. As the training progresses, recall increases steadily, with significant improvements observed during the early to mid-epochs. There is some fluctuation in the recall values, particularly around the middle epochs, as the model adjusts its parameters and learns the underlying patterns in the dataset. Toward the later epochs, the recall stabilizes at higher levels, with a noticeable upward trend near the end, indicating the model's increasing capability to detect positive instances accurately. The overall trend highlights the gradual improvement in the model's recall performance as training advances. This graph provides insights into how the YOLOv7 model's ability to detect objects evolves and improves with each epoch.

\begin{figure}[H]
\begin{adjustwidth}{-\extralength}{0cm}
\centering
\includegraphics[height=8cm]{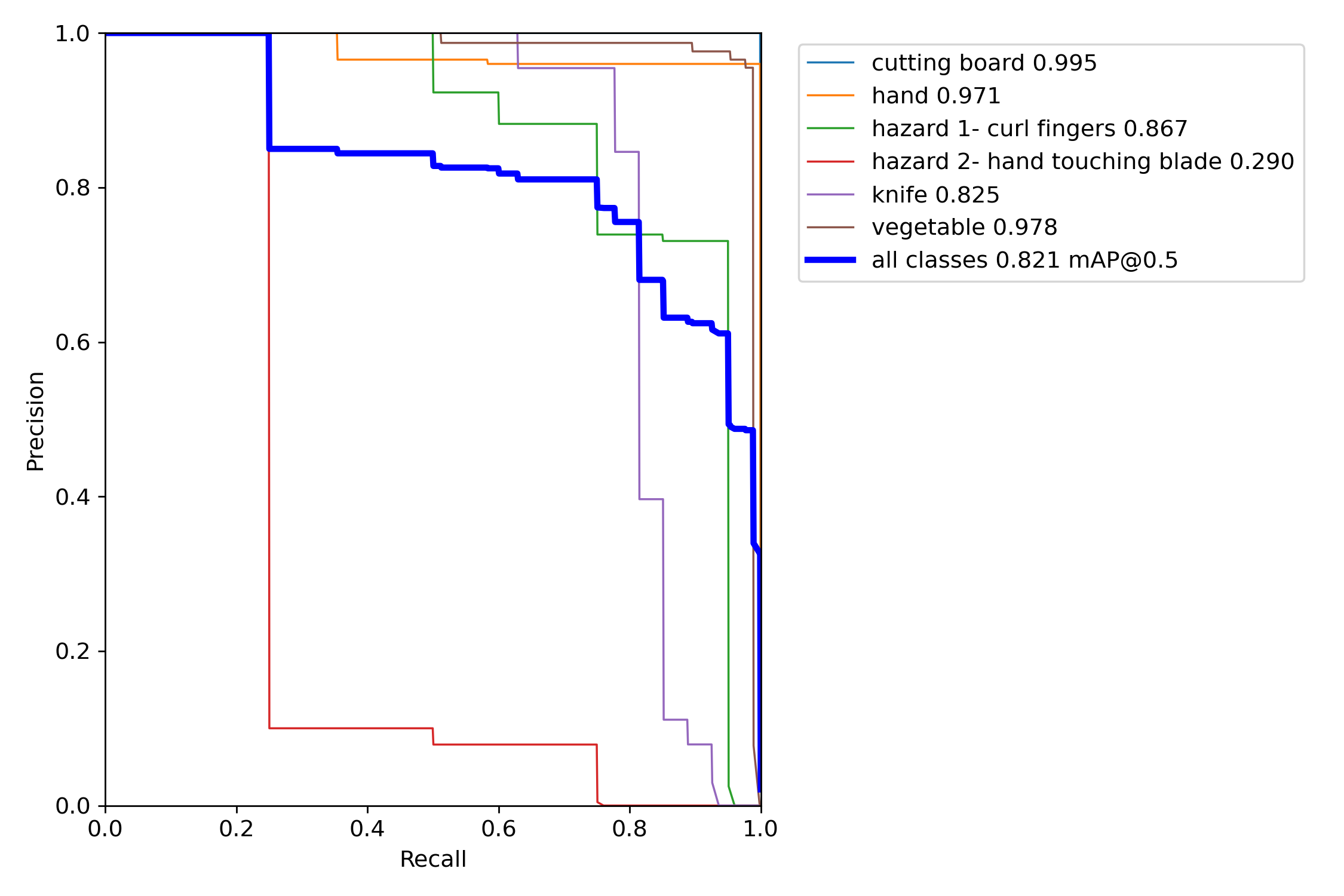}
\end{adjustwidth}
\caption{Precision versus recall of YOLOv7.}
\label{Figure:7}
\end{figure}

In precision-recall curve, the x-axis represents recall, which measures the proportion of true positive detections out of all actual positives, while the y-axis shows precision, the proportion of true positives out of all predicted positives (Figure \ref{Figure:7}). The "cutting board" class achieves the highest AP of 0.995, while "hazard 2 - hand touching blade" has a significantly lower AP of 0.290. A mean average precision (mAP) of 0.821 at an Intersection over Union (IoU) threshold of 0.5 was achieved in the overall model performance across all classes. Classes like "hand" (0.971 AP) and "vegetable" (0.978 AP) demonstrate excellent detection performance, while "hazard 2 - hand touching blade" struggles, reflecting challenges in detecting certain object categories.

\begin{figure}[H]
\begin{adjustwidth}{-\extralength}{0cm}
\centering
\includegraphics[height=7cm]{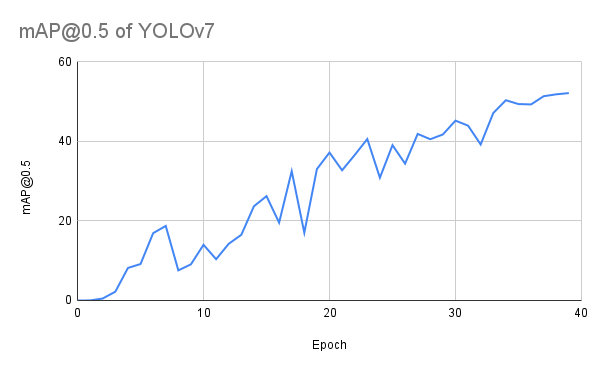}
\end{adjustwidth}
\caption{mAP50 of YOLOv7.}
\label{Figure:8}
\end{figure}

The mean Average Precision (mAP) at an IoU threshold of 0.5 (mAP0.5) for the YOLOv7 model over 40 epochs is shown in Figure \ref{Figure:8}. Initially, the mAP0.5 starts at a very low value, reflecting the model's limited detection capabilities in the early stages of training. As training progresses, the mAP0.5 increases steadily, with fluctuations in the middle epochs as the model fine-tunes its ability to localize and classify objects. The upward trend toward the later epochs demonstrates the model's improvement in performance, with the mAP0.5 stabilizing and reaching higher levels by the end of training. Overall, this graph highlights the YOLOv7 model's gradual learning and improvement in object detection accuracy, achieving a solid performance by the end of the training process.

\begin{figure}[H]
\begin{adjustwidth}{-\extralength}{0cm}
\centering
\includegraphics[height=7cm]{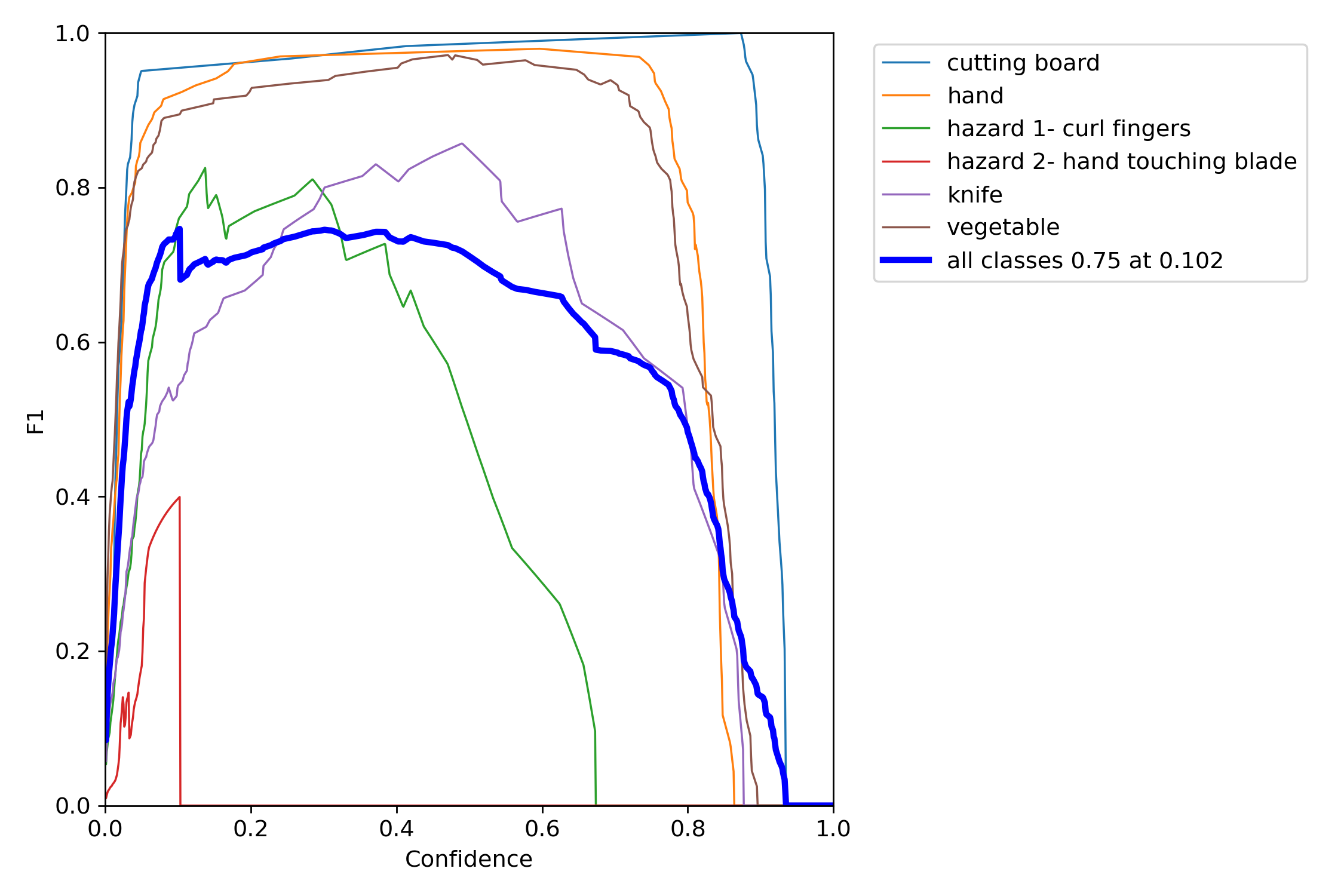}
\end{adjustwidth}
\caption{F1 curve of YOLOv7.}
\label{Figure:9}
\end{figure}

In F1 curve, the value "0.75 at 0.102" in the legend indicates that the overall maximum F1 score is 0.75, achieved at a confidence threshold of 0.102. The curves show that most classes achieve their peak F1 scores at relatively low confidence thresholds before declining as the threshold increases. For example, classes like "cutting board" and "hand" maintain high F1 scores across a wide range of thresholds, reflecting strong precision and recall. In contrast, "hazard 2 - hand touching blade" struggles, with a much lower F1 score, indicating difficulties in accurately detecting this class. Overall, the F1-Confidence curve provides a clear visualization of how the model's performance varies across classes and confidence thresholds, helping to identify an optimal threshold for achieving balanced precision and recall while highlighting classes that may require further optimization (Figure \ref{Figure:9}).

\begin{figure}[H]
\begin{adjustwidth}{-\extralength}{0cm}
\centering
\includegraphics[height=13cm]{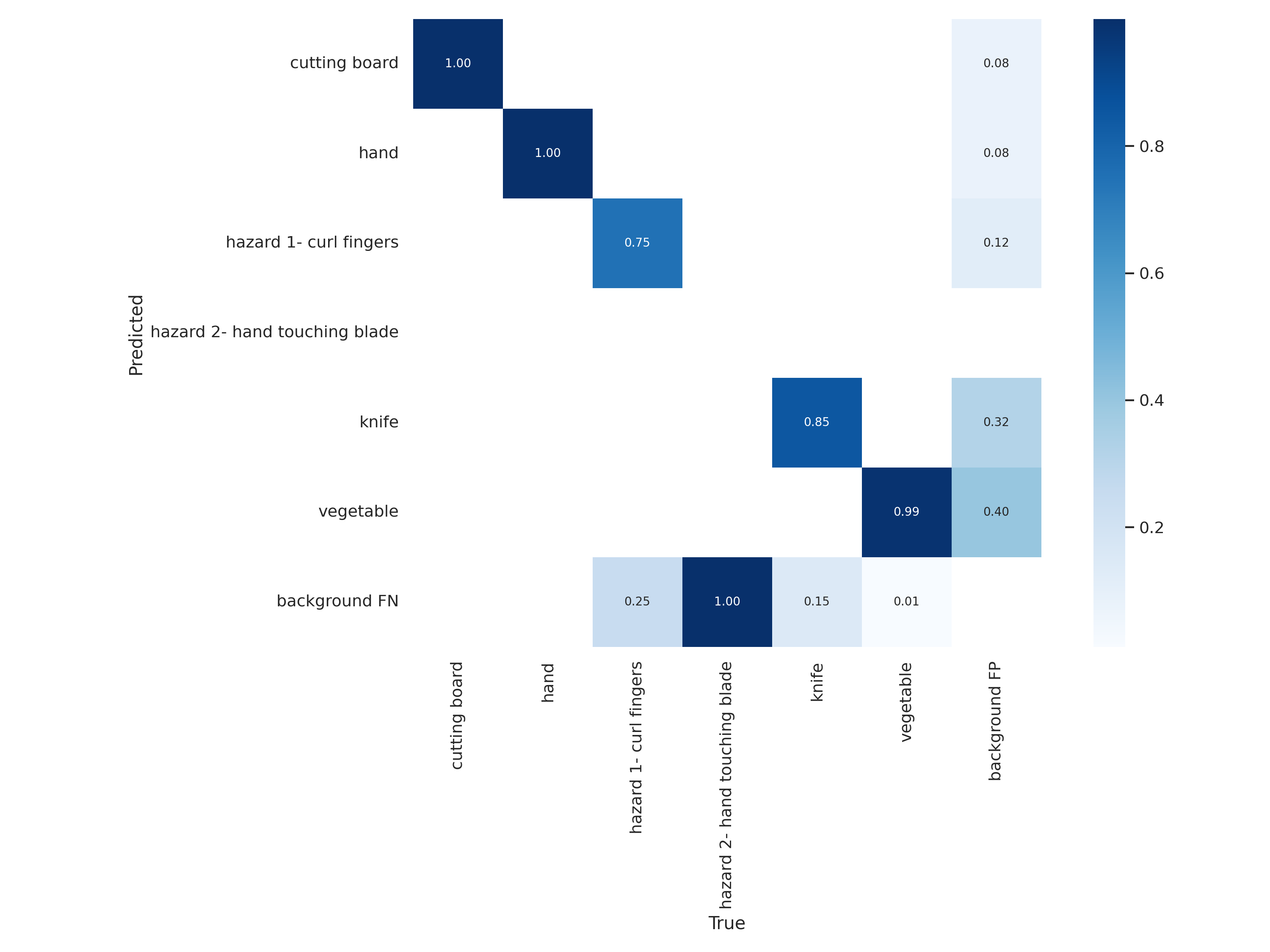}
\end{adjustwidth}
\caption{Confusion matrix of YOLOv7.}
\label{Figure:10}
\end{figure}

Confusion matrix for the YOLOv7 model shows the model's performance in predicting object classes (Figure \ref{Figure:10}). The model achieves perfect accuracy (1.00) for predicting "cutting board" and "hand," while other classes like "knife" (0.85) and "vegetable" (0.99) show high but not best performance. "Knife" is sometimes misclassified as "vegetable" with a value of 0.32, and "hazard 1 - curl fingers" shows significant misclassification as "hand" with a value of 0.12. Background false negatives (FN) and background false positives (FP) cells highlight the model's errors in detecting or misinterpreting objects in the background. The intensity of the blue shading visually represents the magnitude of the values, with darker shades indicating higher accuracy or higher misclassification rates. Overall, this confusion matrix provides insights into the model's strengths and weaknesses, highlighting areas where the model is best (e.g., "cutting board") and where it is not (e.g., distinguishing "knife" from "vegetable").

To conclude, the precision graph shows steady growth over training, stabilizing at higher levels in later epochs. The recall graph similarly indicates consistent improvement, with high levels achieved by the end of training. The mAP@0.5 graph demonstrates a steady increase in overall detection performance, with the model effectively balancing precision and recall by the final epochs. The F1 curve highlights YOLOv7's strong overall performance with a peak F1 score of 0.75 at a 0.102 confidence threshold. The confusion matrix reveals excellent performance for certain classes, such as "cutting board" and "hand," with the best accuracy, but misclassifications for challenging categories like "hazard 1 - curl fingers" and "hazard 2 - hand touching blade." There is also some confusion between "knife" and "vegetable" and issues with background false positives and negatives, suggesting room for improvement. Overall, YOLOv7 demonstrates strong performance, achieving high accuracy and generalization for most classes.

\section{Discussion}

The evaluation of YOLOv7 using various performance metrics and visualization tools provides valuable insights into the strengths and limitations of the model. The precision and recall graphs show an improvement during training, indicating that the model effectively learns to balance the trade-offs between correctly identifying objects and minimizing false positives. The stabilization of both precision and recall in the later epochs suggests that the model achieves a level of understanding of the dataset. Fluctuations observed in the middle epochs reflect the challenges the model faces when learning complex patterns.

The mAP@0.5 metric further reinforces the model's strong performance, with a steady increase throughout the training process. This metric, which balances localization and classification accuracy, highlights the model's effectiveness in detecting objects across a range of confidence thresholds. The F1-Confidence curve further supports these findings by illustrating the model's ability to balance precision and recall across confidence thresholds. The peak F1 score of 0.75 at a 0.102 confidence threshold highlights the model's overall effectiveness.

The confusion matrix provides a better view of the model's behavior, revealing areas where it excels and struggles. Classes such as "cutting board" and "hand" achieve near-perfect accuracy, demonstrating the model's capability to handle well-defined objects. However, misclassifications of classes, such as confusion between "knife" and "vegetable" or difficulties in detecting "hazard 1 - curl fingers" and "hazard 2 - hand touching blade," indicate that the model struggles with visually similar or complex classes.

\section{Conclusions}

This study evaluates the performance of YOLOv7 across multiple metrics, including precision, recall, mAP, F1 score, and confusion matrix analysis, providing a comprehensive understanding of the model’s strengths and limitations. YOLOv7 demonstrates robust object detection capabilities, with high precision and recall achieved during training and a peak mAP@0.5 indicating strong overall accuracy in detecting and localizing objects. The model excels in identifying well-defined classes such as "cutting board" and "hand," but exhibits challenges in distinguishing visually similar or complex classes, such as "knife" and "vegetable," or hazardous scenarios like "hand touching blade." The F1 curve further emphasizes the model’s ability to balance precision and recall, with a peak F1 score of 0.75 at a confidence threshold of 0.102.

Research could be extended to analyze the model's adaptability to different types of knife shapes and food substances, where kitchen tools or equipment obstruct visibility. The dataset can also be enlarged to cover additional hazards and new classes. This investigation could be extended to other sectors, such as renewable energy ~\cite{hussain2019deployment, hussain2022statistical}, healthcare ~\cite{hussain2022exudate} and wireless sensor networks for wide scale edge based deployments ~\cite{alsboui2022dynamic}

\begin{adjustwidth}{-\extralength}{0cm}

\bibliographystyle{unsrt}  
\bibliography{ref}  

\begin{thebibliography}{10}

\bibitem{RN1}
Natthariya Laopracha and Khamron Sunat.
\newblock Comparative study of computational time that hog-based features used for vehicle detection.
\newblock In {\em Recent Advances in Information and Communication Technology 2017: Proceedings of the 13th International Conference on Computing and Information Technology (IC2IT)}, pages 275--284. Springer, 2018.

\bibitem{RN2}
J~Kittler, M~Hatef, Robert~PW Duin, and J~Matas.
\newblock On combining classifiers. ieee transaction on pattern analysis and machine intelligence, 1998.

\bibitem{RN3}
H~Wang, A~Klaser, C~Schmid, and CL~Liu.
\newblock Action recognition by dense trajectories. cvpr’11, washington, dc, usa.
\newblock {\em IEEE Computer Society}, pages 3169--3176, 2011.

\bibitem{RN4}
Marcus Rohrbach, Sikandar Amin, Mykhaylo Andriluka, and Bernt Schiele.
\newblock A database for fine grained activity detection of cooking activities.
\newblock In {\em 2012 IEEE conference on computer vision and pattern recognition}, pages 1194--1201. IEEE, 2012.

\bibitem{RN5}
M~Milagro Fernandez-Carrobles, Oscar Deniz, and Fernando Maroto.
\newblock Gun and knife detection based on faster r-cnn for video surveillance.
\newblock In {\em Iberian conference on pattern recognition and image analysis}, pages 441--452. Springer, 2019.

\bibitem{RN8}
Kaiming He, Xiangyu Zhang, Shaoqing Ren, and Jian Sun.
\newblock Deep residual learning for image recognition.
\newblock In {\em Proceedings of the IEEE conference on computer vision and pattern recognition}, pages 770--778, 2016.

\bibitem{RN7}
Huanhuan Ran, Shiping Wen, Kaibo Shi, and Tingwen Huang.
\newblock Stable and compact design of memristive googlenet neural network.
\newblock {\em Neurocomputing}, 441:52--63, 2021.

\bibitem{RN10}
Ross Girshick, Jeff Donahue, Trevor Darrell, and Jitendra Malik.
\newblock Rich feature hierarchies for accurate object detection and semantic segmentation.
\newblock In {\em Proceedings of the IEEE conference on computer vision and pattern recognition}, pages 580--587, 2014.

\bibitem{RN6}
Alex Krizhevsky, Ilya Sutskever, and Geoffrey~E Hinton.
\newblock Imagenet classification with deep convolutional neural networks.
\newblock {\em Communications of the ACM}, 60(6):84--90, 2017.

\bibitem{RN11}
Mounika Gajja.
\newblock Brain tumor detection using mask r-cnn.
\newblock {\em J. Adv. Res. Dyn. Control Syst}, 12:101--108, 2020.

\bibitem{RN12}
Shuang Liu, Xing Cui, Jiayi Li, Hui Yang, and Niko Luka{\v{c}}.
\newblock Pedestrian detection based on faster r-cnn.
\newblock {\em International Journal of Performability Engineering}, 15(7):1792, 2019.

\bibitem{RN9}
Zihan Yang.
\newblock Classification of picture art style based on vggnet.
\newblock In {\em Journal of Physics: Conference Series}, volume 1774, page 012043. IOP Publishing, 2021.

\bibitem{RN13}
Muhammad Hussain.
\newblock Yolo-v5 variant selection algorithm coupled with representative augmentations for modelling production-based variance in automated lightweight pallet racking inspection.
\newblock {\em Big Data and Cognitive Computing}, 7(2):120, 2023.

\bibitem{RN14}
Joseph Redmon, Santosh Divvala, Ross Girshick, and Ali Farhadi.
\newblock You only look once: Unified, real-time object detection.
\newblock In {\em Proceedings of the IEEE conference on computer vision and pattern recognition}, pages 779--788, 2016.

\bibitem{RN15}
Joseph Redmon and Ali Farhadi.
\newblock Yolo9000: better, faster, stronger.
\newblock In {\em Proceedings of the IEEE conference on computer vision and pattern recognition}, pages 7263--7271, 2017.

\bibitem{RN16}
Joseph Redmon and Ali Farhadi.
\newblock Yolov3: An incremental improvement.
\newblock {\em arXiv preprint arXiv:1804.02767}, 2018.

\bibitem{RN17}
Alexey Bochkovskiy, Chien-Yao Wang, and Hong-Yuan~Mark Liao.
\newblock Yolov4: Optimal speed and accuracy of object detection.
\newblock {\em arXiv preprint arXiv:2004.10934}, 2020.

\bibitem{RN18}
Lakshantha Dissanayake, Glenn Jocher, Q~Burhan, and Sergiu Waxmann.
\newblock Comprehensive guide to ultralytics yolov5, 2023.

\bibitem{RN19}
Chuyi Li, Lulu Li, Hongliang Jiang, Kaiheng Weng, Yifei Geng, Liang Li, Zaidan Ke, Qingyuan Li, Meng Cheng, Weiqiang Nie, et~al.
\newblock Yolov6: A single-stage object detection framework for industrial applications.
\newblock {\em arXiv preprint arXiv:2209.02976}, 2022.

\bibitem{RN20}
Chien-Yao Wang, Alexey Bochkovskiy, and Hong-Yuan~Mark Liao.
\newblock Yolov7: Trainable bag-of-freebies sets new state-of-the-art for real-time object detectors.
\newblock In {\em Proceedings of the IEEE/CVF conference on computer vision and pattern recognition}, pages 7464--7475, 2023.

\bibitem{RN21}
Glenn Jocher, Muhammad~Rizwan Munawar, and Ayush Chaurasia.
\newblock Yolo: A brief history, 2023.

\bibitem{RN22}
Mujadded Al~Rabbani Alif, Sabbir Ahmed, and Muhammad~Abul Hasan.
\newblock Isolated bangla handwritten character recognition with convolutional neural network.
\newblock In {\em 2017 20th International conference of computer and information technology (ICCIT)}, pages 1--6. IEEE, 2017.

\bibitem{RN23}
Cuong Pham and Patrick Olivier.
\newblock Slice\&dice: Recognizing food preparation activities using embedded accelerometers.
\newblock In {\em European Conference on Ambient Intelligence}, pages 34--43. Springer, 2009.

\bibitem{RN24}
Sebastian Stein and Stephen~J McKenna.
\newblock Combining embedded accelerometers with computer vision for recognizing food preparation activities.
\newblock In {\em Proceedings of the 2013 ACM international joint conference on Pervasive and ubiquitous computing}, pages 729--738, 2013.

\bibitem{RN25}
Marek {\.Z}ywicki, Andrzej Matiola{\'n}ski, Tomasz~M Orzechowski, and Andrzej Dziech.
\newblock Knife detection as a subset of object detection approach based on haar cascades.
\newblock In {\em Proceedings of 11th International Conference “Pattern recognition and information processing}, pages 139--142, 2011.

\bibitem{hussain2023custom}
Muhammad Hussain and Richard Hill.
\newblock Custom lightweight convolutional neural network architecture for automated detection of damaged pallet racking in warehousing \& distribution centers.
\newblock {\em IEEE Access}, 11:58879--58889, 2023.

\bibitem{hussain2022feature}
Muhammad Hussain, Hussain Al-Aqrabi, Muhammad Munawar, and Richard Hill.
\newblock Feature mapping for rice leaf defect detection based on a custom convolutional architecture.
\newblock {\em Foods}, 11(23):3914, 2022.

\bibitem{RN26}
Alex Krizhevsky, Ilya Sutskever, and Geoffrey~E Hinton.
\newblock Imagenet classification with deep convolutional neural networks.
\newblock {\em Communications of the ACM}, 60(6):84--90, 2017.

\bibitem{RN40}
Songbo Chen, Wenhu Tang, Tianyao Ji, Huiling Zhu, Ye~Ouyang, and Wenbo Wang.
\newblock Detection of safety helmet wearing based on improved faster r-cnn.
\newblock In {\em 2020 International joint conference on neural networks (IJCNN)}, pages 1--7. IEEE, 2020.

\bibitem{RN27}
David~A Noever and Sam E~Miller Noever.
\newblock Knife and threat detectors.
\newblock {\em arXiv preprint arXiv:2004.03366}, 2020.

\bibitem{RN28}
Samet Akcay, Mikolaj~E Kundegorski, Chris~G Willcocks, and Toby~P Breckon.
\newblock Using deep convolutional neural network architectures for object classification and detection within x-ray baggage security imagery.
\newblock {\em IEEE transactions on information forensics and security}, 13(9):2203--2215, 2018.

\bibitem{RN29}
Van-Hung Le, Hai Vu, and Thuy~Thi Nguyen.
\newblock A frame-work assisting the visually impaired people: common object detection and pose estimation in surrounding environment.
\newblock In {\em 2018 5th NAFOSTED conference on information and computer science (NICS)}, pages 216--221. IEEE, 2018.

\bibitem{RN30}
Joshua van Staden and Dane Brown.
\newblock An evaluation of yolo-based algorithms for hand detection in the kitchen.
\newblock In {\em 2021 International Conference on Artificial Intelligence, Big Data, Computing and Data Communication Systems (icABCD)}, pages 1--7. IEEE, 2021.

\bibitem{RN31}
Qiang Li, Feng Zhao, Zhongping Xu, Kexin Li, Jing Wang, Haofeng Liu, Liang Qin, and Kaipei Liu.
\newblock Improved yolov4 algorithm for safety management of on-site power system work.
\newblock {\em Energy reports}, 8:739--746, 2022.

\bibitem{RN32}
Hubert Ngankam, Philippe Dion, H{\'e}l{\`e}ne Pigot, and Sylvain Giroux.
\newblock Real-time multiple object tracking for safe cooking activities.
\newblock In {\em International Conference on Smart Homes and Health Telematics}, pages 192--204. Springer, 2023.

\bibitem{RN33}
Yunfan Shi, Zheng Yang, Yifei Bi, Jingcheng Li, Xiaohui Zhu, and Yong Yue.
\newblock Realtime mask detection of kitchen staff using yolov5 and edge computing.
\newblock In {\em 2023 3rd International Conference on Computer, Control and Robotics (ICCCR)}, pages 33--40. IEEE, 2023.

\bibitem{RN34}
Isaias Majil, Mau-Tsuen Yang, and Sophia Yang.
\newblock Augmented reality based interactive cooking guide.
\newblock {\em Sensors}, 22(21):8290, 2022.

\bibitem{RN35}
Iker Azurmendi, Ekaitz Zulueta, Jose~Manuel Lopez-Guede, Jon Azkarate, and Manuel Gonz{\'a}lez.
\newblock Cooktop sensing based on a yolo object detection algorithm.
\newblock {\em Sensors}, 23(5):2780, 2023.

\bibitem{RN36}
Saydirasulov Norkobil~Saydirasulovich, Akmalbek Abdusalomov, Muhammad~Kafeel Jamil, Rashid Nasimov, Dinara Kozhamzharova, and Young-Im Cho.
\newblock A yolov6-based improved fire detection approach for smart city environments.
\newblock {\em Sensors}, 23(6):3161, 2023.

\bibitem{RN37}
Yi~Yi Aung and Kyi~Zar Oo.
\newblock Detection of guns and knives images based on yolo v7.
\newblock In {\em 2024 3rd International Conference on Artificial Intelligence For Internet of Things (AIIoT)}, pages 1--6. IEEE, 2024.

\bibitem{RN38}
Maha Mokrani and Zied Hajaiej.
\newblock Real time object detection with data variation.
\newblock {\em Przeglad Elektrotechniczny}, 2024(5), 2024.

\bibitem{RN39}
Anuja Radhakrishnan, Sumisha Samuel, Sachin~Shaju John, Riya~Ann Reji, Stephin John, Liya~Elizabeth Jacob, and Devi Vinod.
\newblock An automated alert system for monitoring the hygiene in restaurants using machine vision.
\newblock In {\em 2024 1st International Conference on Trends in Engineering Systems and Technologies (ICTEST)}, pages 1--6. IEEE, 2024.

\bibitem{RN200}
R~An, X~Zhang, M~Sun, and G~Wang.
\newblock Gc-yolov9: Innovative smart city traffic monitoring solution.
\newblock {\em Alexandria Engineering Journal}, 106:277--287, 2024.

\bibitem{geetha2024comparative}
Athulya~Sundaresan Geetha and Muhammad Hussain.
\newblock A comparative analysis of yolov5, yolov8, and yolov10 in kitchen safety.
\newblock {\em arXiv preprint arXiv:2407.20872}, 2024.

\bibitem{RN204}
Athulya Sundaresan~Geetha, Mujadded Al~Rabbani Alif, Muhammad Hussain, and Paul Allen.
\newblock Comparative analysis of yolov8 and yolov10 in vehicle detection: Performance metrics and model efficacy.
\newblock {\em Vehicles}, 6(3):1364--1382, 2024.

\bibitem{RN205}
Athulya~Sundaresan Geetha.
\newblock Comparing yolov5 variants for vehicle detection: A performance analysis.
\newblock {\em arXiv preprint arXiv:2408.12550}, 2024.

\bibitem{RN206}
Athulya~Sundaresan Geetha.
\newblock What is yolov6? a deep insight into the object detection model.
\newblock {\em arXiv preprint arXiv:2412.13006}, 2024.

\bibitem{hussain2022gradient}
Muhammad Hussain, Tianhua Chen, Sofya Titrenko, Pan Su, and Mufti Mahmud.
\newblock A gradient guided architecture coupled with filter fused representations for micro-crack detection in photovoltaic cell surfaces.
\newblock {\em IEEE Access}, 10:58950--58964, 2022.

\bibitem{aydin2023domain}
Burcu~Ataer Aydin, Muhammad Hussain, Richard Hill, and Hussain Al-Aqrabi.
\newblock Domain modelling for a lightweight convolutional network focused on automated exudate detection in retinal fundus images.
\newblock In {\em 2023 9th International Conference on Information Technology Trends (ITT)}, pages 145--150. IEEE, 2023.

\bibitem{hussain2023child}
Muhammad Hussain and Hussain Al-Aqrabi.
\newblock Child emotion recognition via custom lightweight cnn architecture.
\newblock In {\em Kids Cybersecurity Using Computational Intelligence Techniques}, pages 165--174. Springer, 2023.

\bibitem{RN41}
Mujadded Al~Rabbani Alif and Muhammad Hussain.
\newblock Yolov1 to yolov10: A comprehensive review of yolo variants and their application in the agricultural domain.
\newblock {\em arXiv preprint arXiv:2406.10139}, 2024.

\bibitem{RN201}
Chien-Yao Wang, Alexey Bochkovskiy, and Hong-Yuan~Mark Liao.
\newblock Yolov7: Trainable bag-of-freebies sets new state-of-the-art for real-time object detectors.
\newblock In {\em Proceedings of the IEEE/CVF conference on computer vision and pattern recognition}, pages 7464--7475, 2023.

\bibitem{RN202}
Jacob Solawetz.
\newblock What is yolov7? a complete guide, 2024.

\bibitem{RN203}
Kukil and Sovit Rath.
\newblock Yolov7 object detection paper explanation \& inference, 2022.

\bibitem{hussain2019deployment}
Muhammad Hussain, Mahmoud Dhimish, Violeta Holmes, and Peter Mather.
\newblock Deployment of ai-based rbf network for photovoltaics fault detection procedure.
\newblock {\em AIMS Electronics and Electrical Engineering}, 4(1):1--18, 2019.

\bibitem{hussain2022statistical}
Muhammad Hussain, Hussain Al-Aqrabi, and Richard Hill.
\newblock Statistical analysis and development of an ensemble-based machine learning model for photovoltaic fault detection.
\newblock {\em Energies}, 15(15):5492, 2022.

\bibitem{hussain2022exudate}
Muhammad Hussain, Hussain Al-Aqrabi, Muhammad Munawar, Richard Hill, and Simon Parkinson.
\newblock Exudate regeneration for automated exudate detection in retinal fundus images.
\newblock {\em IEEE access}, 11:83934--83945, 2022.

\bibitem{alsboui2022dynamic}
Tariq Alsboui, Richard Hill, Hussain Al-Aqrabi, Hafiz Muhammad~Athar Farid, Muhammad Riaz, Shamaila Iram, Hafiz~Muhammad Shakeel, and Muhammad Hussain.
\newblock A dynamic multi-mobile agent itinerary planning approach in wireless sensor networks via intuitionistic fuzzy set.
\newblock {\em Sensors}, 22(20):8037, 2022.

\end{thebibliography}

\end{adjustwidth}
\end{document}